# Application of Convolutional Neural Network for Image Classification on Pascal VOC Challenge 2012 dataset

Suyash Shetty
Manipal Institute of Technology
suyash.shashikant@learner.manipal.edu

**Abstract**

In this project we work on creating a model to classify images for the Pascal VOC Challenge 2012. We use convolutional neural networks trained on a single GPU instance provided by Amazon via their cloud service Amazon Web Services (AWS) to classify images in the Pascal VOC 2012 data set. We train multiple convolutional neural network models and finally settle on the best model which produced a validation accuracy of 85.6% and a testing accuracy of 85.24%.

**1. Introduction**

Using convolutional neural networks (CNN) for image classification has become the de facto standard largely due to the success of *Krizhevsky et al.* [1], *Szegedy et al.* [2], *Simonyan and Zisserman* [3] and especially *He et al.* [4], which is now the state-of-the-art architecture for image classification. By stacking convolutional layers on top of each other, amongst other architectural artifacts, highly accurate models for task of image classification, detection and segmentation have been discover. These models have been able to nearly match, if not exceed human performance on certain datasets.

CNNs are not without their own shortcomings though. Due to the sheer size of networks and the millions of parameters to be optimized, the reliance of high-performance systems increases dramatically. Though modern GPU-based systems are capable of perfoming the intensive computations required by a convolutional neural network, implementing a CNN is unsuitable for those without access to such a high-performance system. Also, more complex architectures such as those by [4] take up to 2-3 weeks to train which demonstrate the inherent difficulties in training deep convolutional neural networks.

However, given the fact that CNNs provide the best results compared to other image classification models and also motivated by the credibility of the performance of CNNs by [1]–[4], we train multiple convolutional neural networks of varying depths to the Pascal VOC 2012 dataset. [5]

**2. Dataset**

The Pascal VOC 2012 has a dataset of around 15,000 labeled images belonging to 20 categories. The categories are some commonly found objects like 'cat', 'dog', 'car' etc. The images vary in their dimensions and were downsampled to a fixed resolution of 128x128. Originally, the dataset was split 50:50 for the training and validation set but we split the data 60:40 in the favour of training data. Each category had a varying number of images but each category had at least 500 images for the models to be trained and validated upon. The only pre-processing performed on the images was mean subtraction, so images input to the network were raw mean-centered RGB pixel values for each image.

The Pascal VOC 2012 dataset contains images with multiple labels mapping to a single image but we only focused on classifying the images to a single label. Images in the training set with multiple labels were reduced to a single label arbitrarily which may have hurt our model. In Figure 1, example images containing multiple labels is shown. Figure 1(a) contains a 'train' with a 'person' inside it and Figure 1(b) contains a 'car' in the foreground and a 'bicycle' along with more cars in the background. Our model was trained by considering only a single label due to which important information about other classes may have been lost.

**3. The Model**

The space of possible convolutional neural network architectures is vast, with decision-making responsibilities at multiple junctions. Selecting suitable paramters is a task of utmost importance. But due the large permutations of the possible parameter combinations, this becomes difficult. We consider 5 CNN models by selecting the parameters of our models heuristically and describe them in this section.

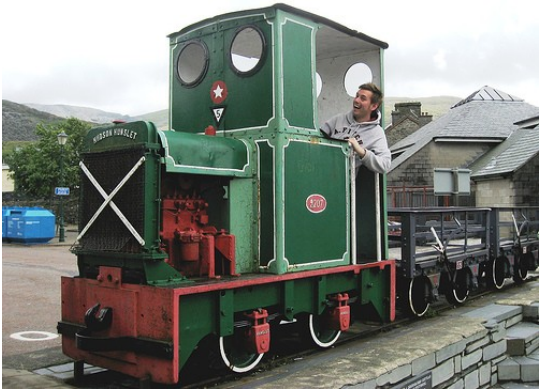 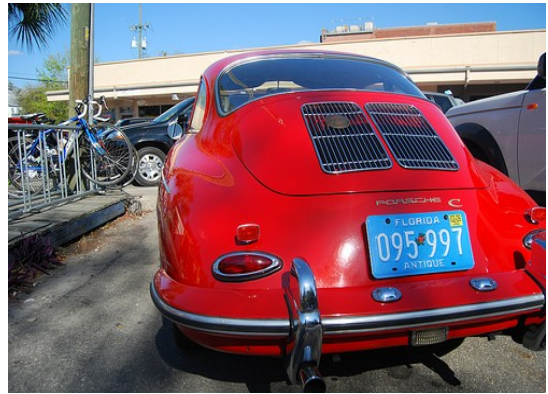

(a)          (b)

Figure 1: Some examples of images in the data set that were originally multi-labeled. (a) was labeled as 'train' and 'person' while (b) was labeled as 'car' and 'bicycle'. Both images contain multiple labels but our model will only classify the most dominant object which will be 'train' in (a) and 'car' in (b).

### 3.1 Network Architecture

We trained 5 Convolutional Neural Networks that vary in their depth, filter sizes and the types of layers. The architecture of the models and their accuracy on the validation set are given in Table 1.

#### 3.1.1 Model 1

Model 1 (M1) is the simplest architecture amongst the five, containing just a single convolution layer and one fully connected layer. The purpose of M1 is to set a benchmark which all following models should surpass. It was encouraging to note that such a simple architecture was able to correctly classify images with a validation accuracy of nealy 62%.

#### 3.1.2 Model 2

Model 2 (M2) builds on M1 and has more convolution layers than its predecessor. It makes use of three convolution layers stacked on top of each other followed by one fully connected layer. Max-pooling layers were placed at certain locations in the architecture to reduce the number of paramters and computation in the network. This along with dropout layers (explained in section 3.3.2) were used to control overfitting.

#### 3.1.3 Model 3

Model 3 (M3) is an exact replica of M2 with the sole exception being the presence of Local Response Normalization (LRN) layers, as used by *Krizhevsky et al.*[1]. These LRN layers resulted in the validation accuracy going up by 3.1%.

#### 3.1.4 Model 4

Model 4 (M4), along with Model 5 (M5) has most complex architecture with 5 convolution layers, 2 pooling layers, 5 LRN layers and 2 fully connected layers. In terms of complexity, it is almost as complex as the Alexnet architecture [1]. It was modelled with the expectation of outperforming all previous models but failed to do so due to either it's slow convergence or due to the naive selection of the number of filters in the convolution layers.

#### 3.1.5 Model 5

Model 5 (M5) has an Alexnet architecture [1] with its weights pre-trained to the Imagenet [6] dataset. The weights for the fully connected layers were randomly initialized with $W \sim N(0, 0.005)$ and the biases were initialized to 0. The model was trained by freezing all layers except the fully connected layers. Unfreezing the convolution layers lead to slightly inferior performance. This model produced the best results with a validation accuracy of 85.6%.

### 3.2 ReLU Non-Linearity

There are several non-linear activation functions that can be used as the output of layer of each convolution or a fully-connected layer such as the sigmoid function and tanh function. But recently, following the success of *Nair and Hinton*[7], using the rectified linear unit or ReLU has become the norm. The ReLU function computes $f(x) = max(0, x)$. The decision to use the ReLU non-linearity also comes from the assertion by [1], that CNNs with ReLU acitvation units train several times faster than equivalent CNNs with tanh activation units.

| Name | Architecture | Validation Accuracy |
|------|--------------|---------------------|
| M1 | IMG-(Conv64-ReLU)-(FC1024-ReLU-FC20)-Softmax | 61.8 % |
| M2 | IMG-(Conv64-ReLU-MaxPool)-(Conv128-ReLU)-(Conv256-ReLU-MaxPool)-(FC1024-ReLU-Dropout-FC20)-Softmax | 71.6 % |
| M3 | IMG-(Conv64-ReLU-LRN-MaxPool)-(Conv128-ReLU-LRN)-(Conv256-ReLU-MaxPool-Dropout)-(FC1024-ReLU-Dropout-FC20)-Softmax | 74.7 % |
| M4 | IMG-(Conv64-ReLU-LRN)x2-MaxPool-(Conv96-ReLU-LRN)x3-MaxPool-(FC1024-ReLU-Dropout)x2-FC20-Softmax | 71.4 % |
| M5 | Pre-trained AlexNet with fine-tuned FC layers | **85.6 %** |

Table 1: The five convolutional neural network models and their corresponding validation accuracies. The number given after the layer name denotes the number of filters in the case of a convolution layer and the number of neurons in the case of a fully-connected layer.

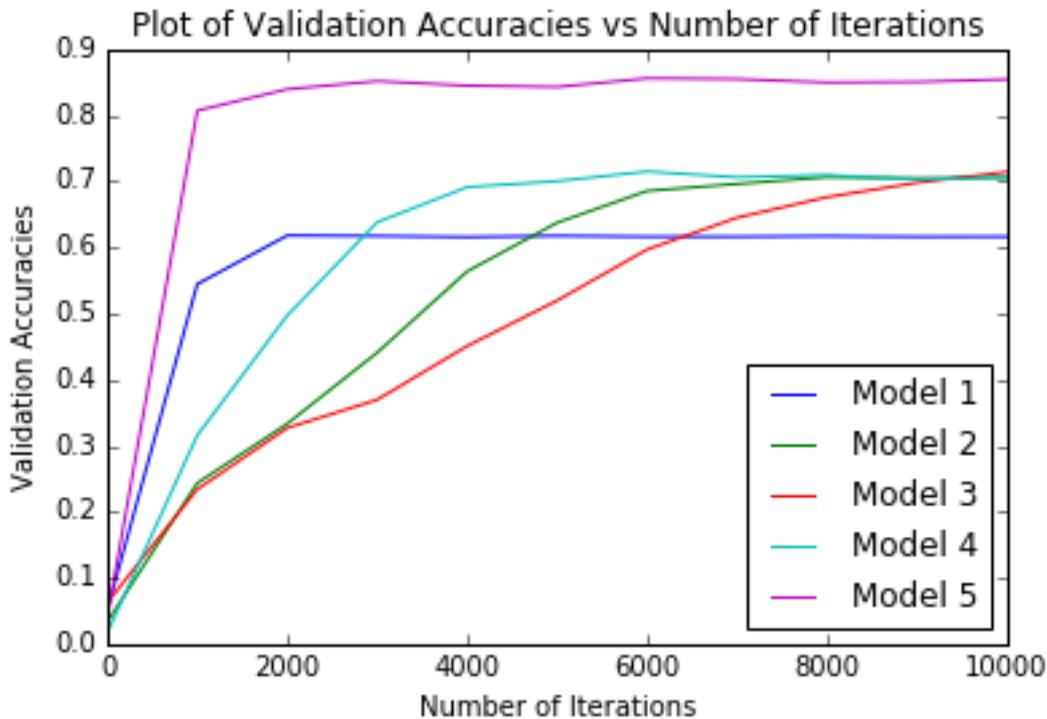

Figure 2: Variation of validation accuracies of each model with the number of iterations. Model 5 has the highest validation accuracy.

### 3.3 Reducing Overfitting

Due to small size of our dataset compared to larger datasets such as CIFAR-10 [8] and ImageNet [6] which have 60,000 images and over a million images respectively, each of our initial four models exhibited large amounts of overfitting. Below we describe the two ways to reduce overfitting. Results of employing the following methods are shown in Figure 3.

### 3.3.1 Data Augmentation

The most basic way to reduce overfitting on image data is to artificially enlarge the training data set using label-preserving transformations. This can be done in a number of ways of which only two were employed here.

One way of augmentating the data set is to horizontally flip an image to produce a new image. A mirror image will have the same label has the original image and hence will preserve the label. Another way is to crop a fixed-sized region from a random point in the image. Taking three such crops for each image and it's horizontal reflection increases our dataset by a factor of 5. Other forms of data augmentation involve performing rotations, varying the intensity or contrast and performing PCA over the images which have proven to further reduce the error rate.

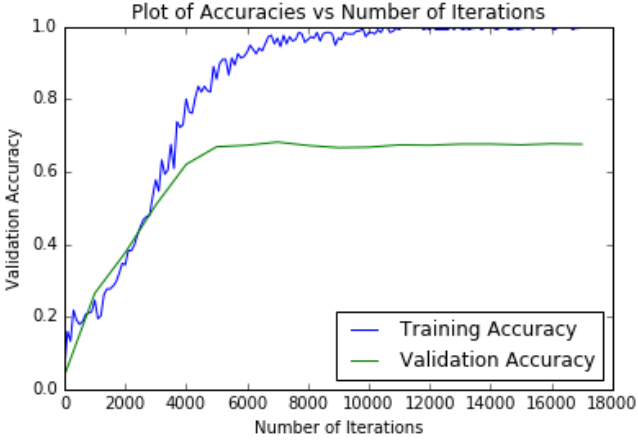 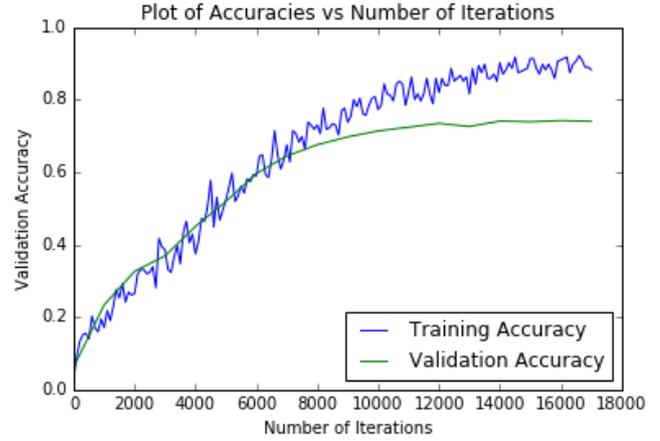

(a)                  (b)

Figure 3: (a) and (b) show variation of the raining and validation accuracy with the number of iterations for M3. 2(a) exhibits substantial overfitting which is evidenced by the validation accuracy saturating at a lower accuracy value. 2(b) shows the result of using dropout and artificially augmenting the data.

### 3.3.2 Regularization

Overfitting was futher reduced by penalizing the squared magnitude of all parameters by adding the term $\lambda \|W\|^2$ to the loss function to each layer, where λ is the hyperparameter that determines the regularization strength and $\|W\|^2$ is the L2-norm of the weights of the layer.

Another technique, called Dropout, introduced by *Srivastava et al.* [9] consists of setting to zero the output of each hidden neuron with a probability p, usually 0.5. These neurons do not participate in the forward and the backward propagation. Therefore, every time an input is presented, the neural network samples a different architecture, while sharing the same weights. At test time, all the neurons are used but their outputs are multiplied by p.

Dropout significantly reduces overfitting but takes more iterations to converge to the local minima. It was also noted that adding dropout to convolution layers resulted in inferior performance compared to when it was added to only the fully connected layers.

## 4. Training
### 4.1 Initialization

Models 1-4 were initialized using the 'Xavier' initialization by *Xavier and Bengio*[10] for the weights and the biases were intialized with 0. Model 5 was initialized with a pre-trained Alexnet model from Caffe's Model Zoo.

### 4.2 Optimization

The models were optimized using a gradient-based optimization method which includes an adaptive moment estimation referred to as 'Adam' [11]. The update formula is given by:

$$(m_t)_i = \beta_1 (m_t-1)_i + (1-\beta_1)(\nabla L(W_t))_i,$$
$$(v_t)_i = \beta_2 (v_t-1)_i + (1-\beta_2)(\nabla L(W_t))_i^2$$

and,

$$(W_{t+1})_i = (W_t)_i - \alpha \frac{\sqrt{1-(\beta_2)_i^t}}{1-(\beta_1)_i^t} \frac{(m_t)_i}{\sqrt{(v_t)_i}+\varepsilon}$$

with $\beta_1 = 0.9, \beta_2 = 0.999, \varepsilon = 10^{-8}$

The learning rate was initialized to $\alpha = 10^{-4}$ and the learning process was manually monitored by lowering the value of α by a factor of 10 when the validation accuracy or training loss saturated to a fixed value. The models were trained for roughly 10,000-20,000 iterations.

### 4.3 Implementation

All models were implemented using Caffe [12], an open-source deep learning framework for convolutional neural networks. All training took on a g2.2xlarge EC2 instance on AWS. This was not an optimal setup and the computational resources often stymied the project due to sudden downtime of spot instances amongst other issues. The memory of the GPU of the instance was limited too and hence made it difficult to train CNNs with larger batch sizes and large number of paramters. Access to better computing resources would yield in much better accuracy due to possibility of training larger models,

## 5. Discussion

This project brings to light many of subtleties of convolutional neural networks and the heuristic approach of selecting a suitable model. M1, with only one convolution layer and one fully connected layer does a substandard job of correctly classifying images. As we increase the complexity of our network architecture and increase its depth, the models perform better, as is expected. However, the importance of selecting appropriate parameters is also brought to light by M4, which is deeper and more complex than M3 but performs significantly worse. As a result, the assertion that deeper CNNs perform better than shallow networks is generally applicable, but cannot be blindly assumed to be true.

We also see the power of supplementary layers like dropout and LRN which improve performance of our models. Dropout sacrifices training accuracy since the entire network is not used during training, but improves validation accuracy to due to the added noise as a result of randomly dropping neurons from a particular layer. The effectiveness of dropout can be confirmed by our work which shows a small increase in validation accuracy in Figure 2. The overuse of dropout however, resulted in degradation of performance. A network with architecture where a dropout layer followed each convolution and fully connected layer showed very poor performance, in addition to the extremely slow convergence rate.

Increased depth, dropout, LRN improve network performance but still fall short compared to transfer learning. M5, which is a pre-trained Alexnet model, performs far better than all our other models. That is not to suggest that transfer learning will always outperform other models but simply to demonstrate its usefullness in certain situations. M5 was pre-trained on an Imagenet dataset which contains similar images but more than one million images. In comparison, our dataset is nearly $(\frac{1}{80})^{th}$ of the dataset that M5 was originally trained. As a result, by using the convolution layers of Alexnet with their pre-trained weights as a CNN feature extractor and then finetuning the newly connected fully connected layers, we achieve much better results.

Some ideas for improving the performance of our models include transfer learning multiple models using different pre-trained weights such as [3], [13] and using a model ensemble of these models. Model ensembles have proven to improve accuracy. Also, convolutional neural networks with deeper layers, more data augmentation, implementing Batch Normalization[14] can prove to be useful for improving the performance.

## References

[1] A. Krizhevsky, I. Sutskever, and G. E. Hinton, "ImageNet Classification with Deep Convolutional Neural Networks," in Advances in Neural Information Processing Systems 25, F. Pereira, C. J. C. Burges, L. Bottou, and K. Q. Weinberger, Eds. Curran Associates, Inc., 2012, pp. 1097–1105.

[2] C. Szegedy, W. Liu, Y. Jia, P. Sermanet, S. Reed, D. Anguelov, D. Erhan, V. Vanhoucke, and A. Rabinovich, "Going Deeper with Convolutions," ArXiv14094842 Cs, Sep. 2014.

[3] K. Simonyan and A. Zisserman, "Very Deep Convolutional Networks for Large-Scale Image Recognition," ArXiv14091556 Cs, Sep. 2014.

[4] K. He, X. Zhang, S. Ren, and J. Sun, "Deep Residual Learning for Image Recognition," ArXiv151203385 Cs, Dec. 2015.

[5] M. Everingham, L. Van Gool, C. K. I. Williams, J. Winn, and A. Zisserman, The PASCAL Visual Object Classes Challenge 2012 (VOC2012) Results.

[6] J. Deng, W. Dong, R. Socher, L.-J. Li, K. Li, and L. Fei-Fei, "ImageNet: A Large-Scale Hierarchical Image Database," in CVPR09, 2009.

[7] V. Nair and G. E. Hinton, "Rectified Linear Units Improve Restricted Boltzmann Machines," presented at the Proceedings of the 27th International Conference on Machine Learning (ICML-10), 2010, pp. 807–814.

[8] A. Krizhevsky, "Learning Multiple Layers of Features from Tiny Images," 2009.

[9] N. Srivastava, G. Hinton, A. Krizhevsky, I. Sutskever, and R. Salakhutdinov, "Dropout: A Simple Way to Prevent Neural Networks from Overfitting," J Mach Learn Res, vol. 15, no. 1, pp. 1929–1958, Jan. 2014.

[10] X. Glorot and Y. Bengio, "Understanding the difficulty of training deep feedforward neural networks," in In Proceedings of the International Conference on Artificial Intelligence and Statistics (AISTATS'10). Society for Artificial Intelligence and Statistics, 2010.

[11] D. Kingma and J. Ba, "Adam: A Method for Stochastic Optimization," ArXiv14126980 Cs, Dec. 2014.

[12] Y. Jia, E. Shelhamer, J. Donahue, S. Karayev, J. Long, R. Girshick, S. Guadarrama, and T. Darrell, "Caffe: Convolutional Architecture for Fast Feature Embedding," ArXiv Prepr. ArXiv14085093, 2014.

[13] P. Sermanet, D. Eigen, X. Zhang, M. Mathieu, R. Fergus, and Y. LeCun, "OverFeat: Integrated Recognition, Localization and Detection using Convolutional Networks," ArXiv13126229 Cs, Dec. 2013.

[14] S. Ioffe and C. Szegedy, "Batch Normalization: Accelerating Deep Network Training by Reducing Internal Covariate Shift," ArXiv150203167 Cs, Feb. 2015.

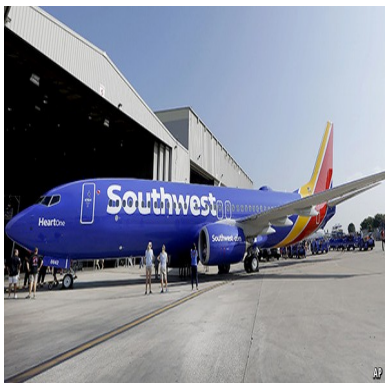 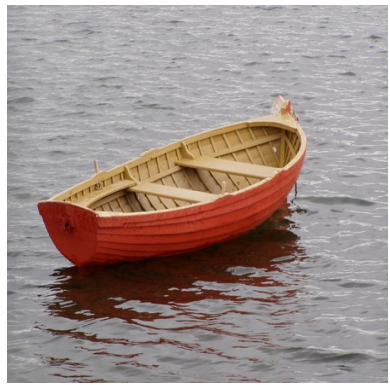 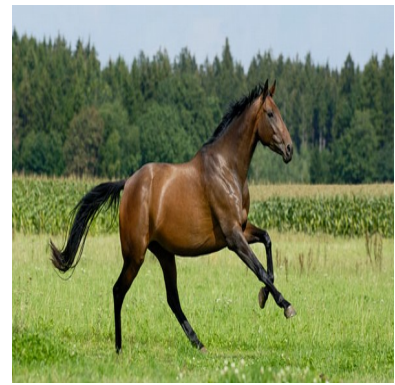

(a)     (b)     (c)

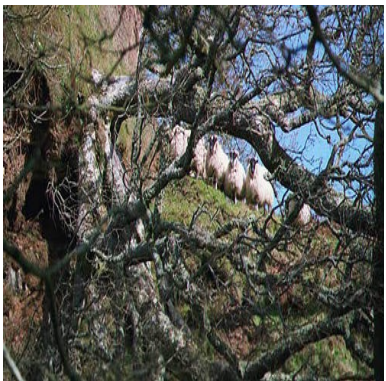 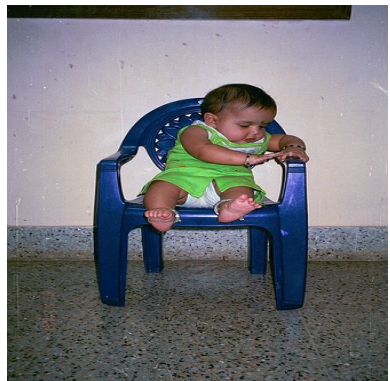 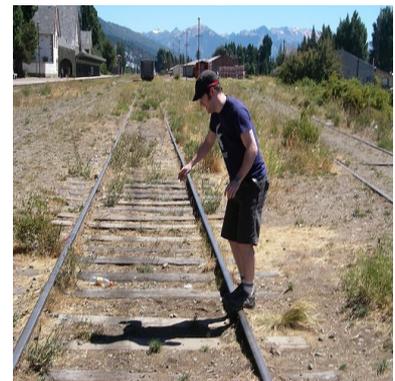

(d)     (e)     (f)

Figure 3: Top row shows all the correct the predictions made by M5. Bottom row shows all the incorrect predictions made by M5. (a), (b) and (c) were correctly classified as 'aeroplane', 'boat' and 'horse' respectively. Whereas, (d), (e) and (f) were incorrectly classified as 'bird', 'car' and 'horse'.